\documentclass[conference]{IEEEtran}
\usepackage{amsmath,graphicx}
\usepackage{amsmath}
\usepackage{amssymb}
\usepackage{booktabs}
\usepackage{multirow}
\usepackage[ruled,lined,linesnumbered]{algorithm2e}
\usepackage{balance}
\usepackage{subfigure}
\usepackage{balance}
\usepackage[table]{xcolor}

\ifCLASSINFOpdf
\else
\fi
\begin{document}

\title{Latent HyperNet: Exploring the Layers \\ of 
Convolutional Neural Networks}

\author{\IEEEauthorblockN{Artur Jordao,
Ricardo Kloss,
William Robson Schwartz}
\IEEEauthorblockA{Smart Sense Laboratory, Computer Science Department\\
Universidade Federal de Minas Gerais, Brazil\\ 
Email: \{arturjordao, rbk, william\}@dcc.ufmg.br}
}

\maketitle
\begin{abstract}
Since Convolutional Neural Networks (ConvNets) are able to simultaneously learn features and classifiers to discriminate different categories of activities, recent works have employed ConvNets approaches to perform human activity recognition (HAR) based on wearable sensors, allowing the removal of expensive human work and expert knowledge. However, these approaches have their power of discrimination limited mainly by the large number of parameters that compose the network and the reduced number of samples available for training. Inspired by this, we propose an accurate and robust approach, referred to as \emph{Latent HyperNet} (LHN). The LHN uses feature maps from early layers (hyper) and projects them, individually, onto a low dimensionality (latent) space. Then, these latent features are concatenated and presented to a classifier. To demonstrate the robustness and accuracy of the LHN, we evaluate it using four different network architectures in five publicly available HAR datasets based on wearable sensors, which vary in the sampling rate and number of activities.	We experimentally demonstrate that the proposed LHN is able to capture rich information, improving the results regarding the original ConvNets. Furthermore, the method outperforms existing state-of-the-art methods, on average, by 5.1 percentage points. 
\end{abstract}
\section{Introduction}\label{sec:introduction}
Human activity recognition (HAR) has received great attention in the past decade since it is fundamental to healthcare, homeland security and smart environments applications. In particular, human activity recognition based on wearable sensors has attracted the attention of the research community mainly due to easy acquisition and processing of the data~\cite{Mukhopadhyay:2015, Yin:2016,Karagiannaki:2017}. 

Recent technological advances have allowed this task to migrate from dedicated wearable sensors to sophisticated devices such as smartphones and smartwatches. Besides, these advances also have enabled the use of different sensors (e.g., accelerometer, gyroscope and barometer), which allow performing an improved activity recognition. However, HAR based on wearable sensors faces a large number of challenges, for instance, treatment of noise and definition of discriminative features able to distinguish the different categories of activities~\cite{Shoaib:2015, Bruno:2015}. 

Many works have demonstrated that the feature extraction process is the most important step regarding HAR based on wearable sensors, since finding adequate features significantly improves the activity recognition rate~\cite{Catal:2015, Shoaib:2015, Karagiannaki:2017}. In the past decade, handcrafted features, such as average, standard deviation and Fourier-based descriptors, were employed to extract higher level descriptions from raw signal data to be presented to a classifier. However, this paradigm requires expert knowledge and expensive human work. 

Recent works employ Convolutional Neural Network (ConvNets) to learn  features and the classifier simultaneously. These approaches have achieved better results than works based on handcrafted features and $1$D convolutions~\cite{Chen:2015,Jiang:2015,Ha:2016,Wang:2017}. On the other hand, ConvNets have some delicate points, such as the need of a large number of training samples, the sensibility to unbalanced data and the large number of parameters to be estimated. As a consequence, the accuracy achieved by ConvNets might be compromised.

\begin{figure}[!t]
	\centering
	\subfigure[Projection using the feature maps at the last layer.] {\includegraphics[scale=0.35]{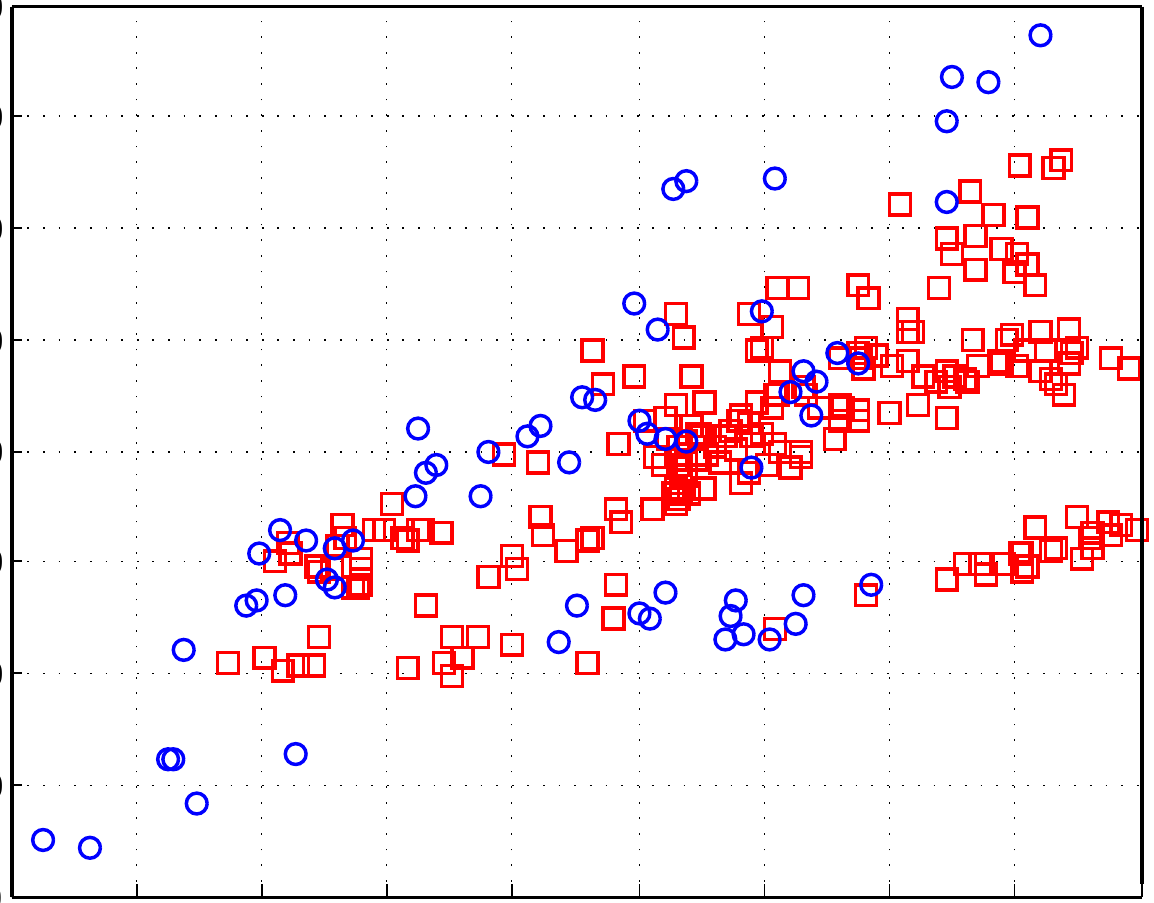}} 
	\hspace{3pt}
	\subfigure[Projection using the features maps from all layers.] {\includegraphics[scale=0.347]{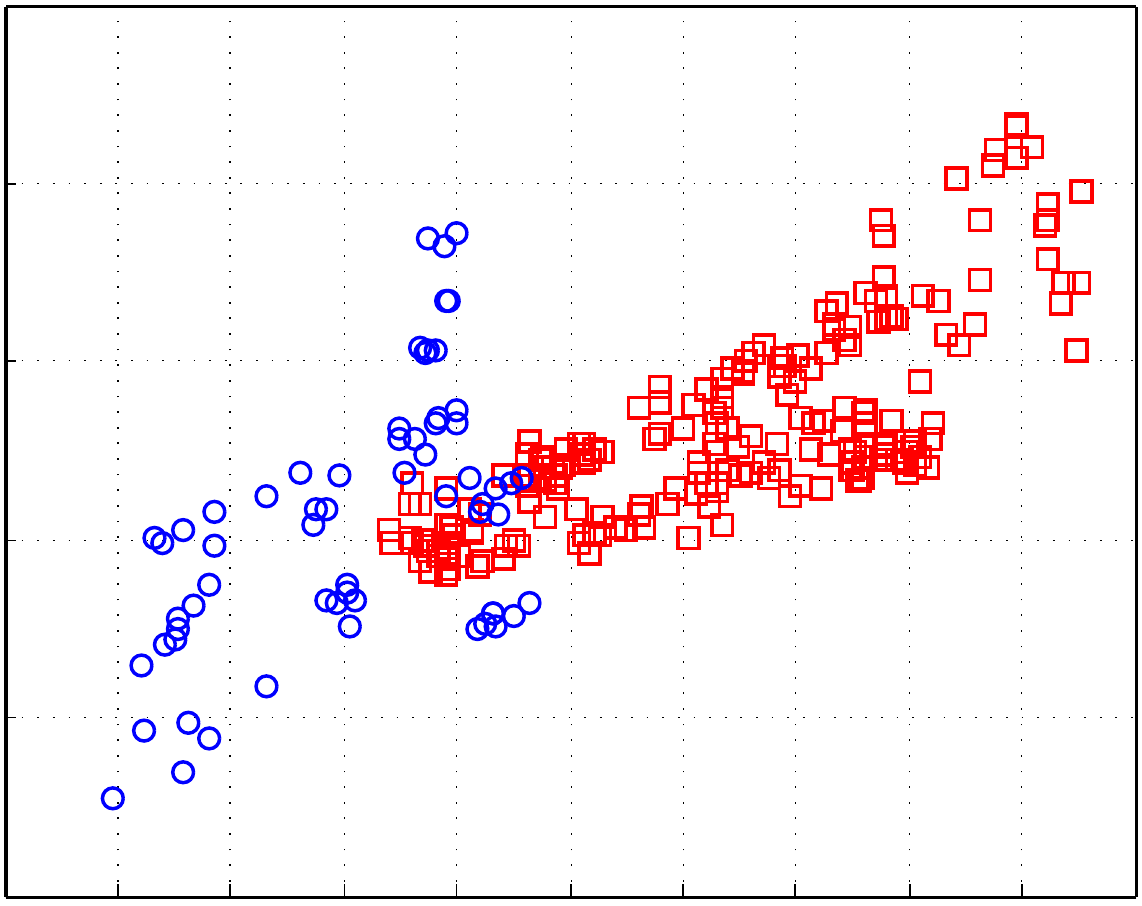}}
	\caption{Projection of two activities onto the two first components of the Partial Least Squares using our ConvNet1 (best viewed in color). {The feature space is better separated when features from early and the last layers are combined. This happens due to multi-scale information provided by the low-level information (shallow layers) with the refined information (deep layers).}}
	\label{fig:projectionFeatureSpace}
\end{figure}

To explore the advantages while facing the drawbacks of the aforementioned issues, in this work we propose an accurate and robust approach, referred to as \emph{Latent HyperNet} (LHN). The LHN relies on the hypothesis that early layers composing a ConvNet provide strong, or complementary, clues to better discriminate the categories of activities. In other words, the combination of low-level information, i.e., shallow layers, with the refined information, i.e., deep layers, might help to better distinguish the activities. Figure~\ref{fig:projectionFeatureSpace} illustrates our hypothesis, showing that the feature space is better separated when features from early layers are combined with features from the last layer, as seen in Figure~\ref{fig:projectionFeatureSpace}(b). 

Similar ideas were employed by Kong et al.~\cite{Kong:2016} in the context of object detection, where the authors incorporated features from early layers and employed them jointly to learn a classifier, instead of using only the last convolutional layer (which is the traditional approach). However, because features from earlier layers have a high dimensional space, the computational cost and the number of parameters increased significantly, making its use impracticable in wearable systems due to memory constraints. On the other hand, our proposed latent hypernet explores, iteratively, all the layers (in this work the max-pooling layers) that compose a ConvNet in an efficient way, enabling us to extract richer information to improve recognition rates for HAR associated with wearable sensors.
\begin{figure}[!t]
	\centering
	\includegraphics[scale=0.43]{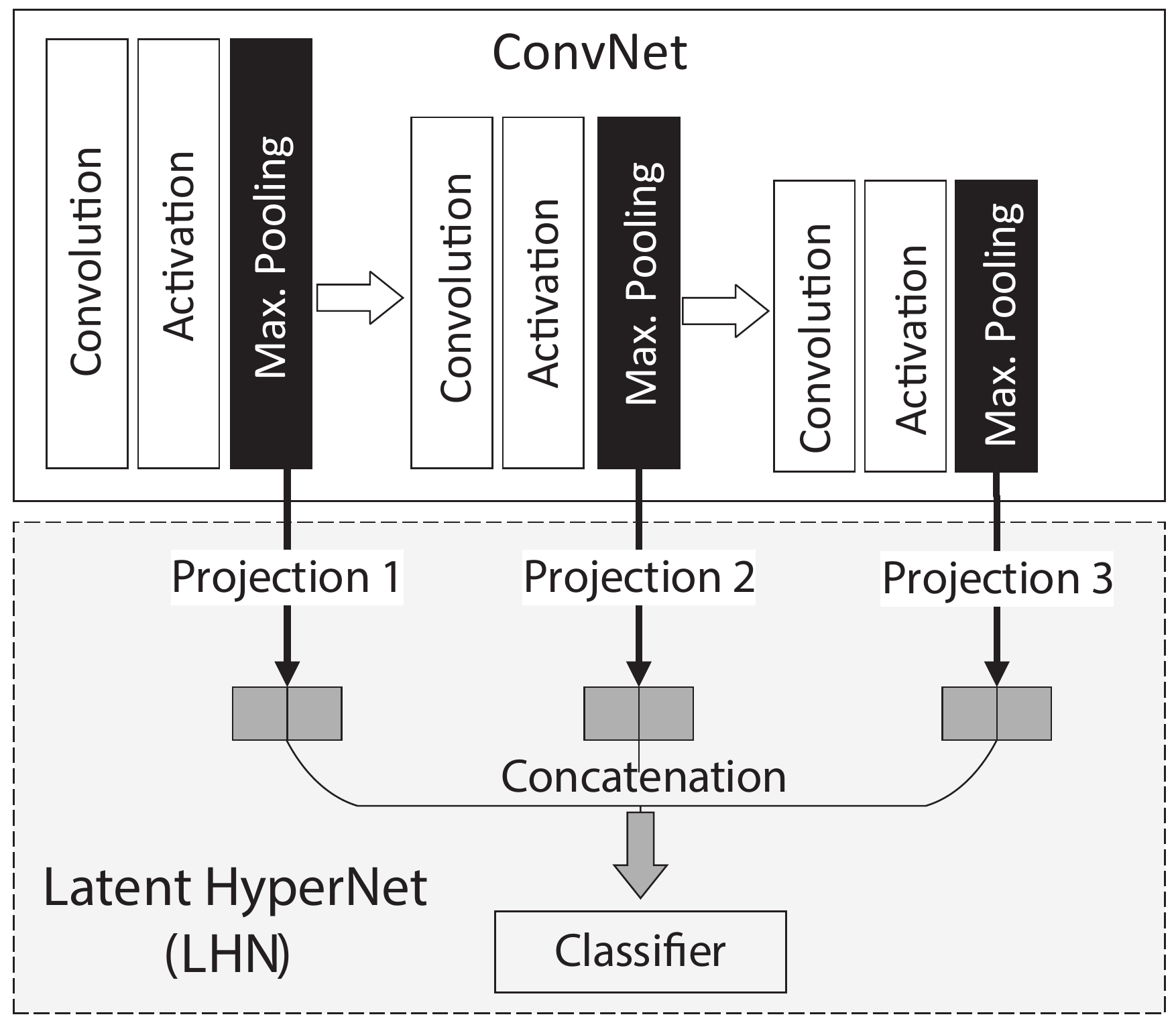}
	\caption{Process to build the Latent HyperNet. After each max-pooling layer, we apply a dimensionality reduction technique to project the features onto a low dimensional space. For this purpose, we use the Partial Least Squares. Then, we concatenate and present the features to a classifier.
	}
	\label{fig:LatentHyperNet}
\end{figure}

The proposed method consists of extracting and projecting features of each layer, individually, onto a latent space using Partial Least Squares~\cite{Wold:1985,Abdi:2010}, {a dimensionality reduction and regression technique widely employed in social sciences and chemometrics to predict a set of dependent variables from a (large) set of independent variables~\cite{Abdi:2010}.} Then, we concatenate and present these features (in the latent space) to a classifier. Figure~\ref{fig:LatentHyperNet} illustrates the LHN approach. It is important to emphasize that we neither modify the design nor the learned weights of the ConvNet during the process to build the LHN. This enables the method to be easily adaptable to any network.

The development of this work presents the following contributions: ($1$) three accurate ConvNets architectures that explore the signal content in different ways, outperforming previous ConvNet architectures specific to HAR based on wearable sensors and serving as insight to future works, with the intention to build ConvNets architectures; ($2$) evidences that early layers that compose a ConvNet provide discriminative information that can increase the recognition rate when properly combined; ($3$) a novel method, the Latent HyperNet (LHN), that effectively combines the network layers, {without the requirement of re-training the network.}

To validate the robustness of the proposed method regarding the employed ConvNet, we evaluate it on three ConvNets architectures proposed in this work and in a ConvNet proposed by Chen and Xue~\cite{Chen:2015} to HAR based on wearable sensors. We evaluate the recognition rate achieved by the LHN method using five publicly available HAR datasets and compare it with state-of-the-art methods, which are ConvNet architectures built for tackling the problem of wearable data. 

Our experiments show that the proposed LHN is able to capture rich information which improves the results regarding the original ConvNet without compromising the computational cost at the prediction stage. To the best of our knowledge, the LHN achieves notable enhancements, since many efforts have been done to achieve smaller improvements in human activity recognition based on wearable sensor data~\cite{Jiang:2015,Ha:2016}. Furthermore, the proposed approach outperforms existing state-of-the-art methods. {On the datasets USCHAD and WISDM, our method achieves a recognition rate of $83.8\%$ and $88.0\%$, respectively.  Additionally, on the datasets UTD-MHAD1 and UTD-MHAD2, our method is able to obtain a recognition rate of $50.1\%$, $75.3\%$, respectively. These results outperform existing state-of-the-art methods in $5.1$ percentage points, on average.}

The remainder  of this paper is structured as follows. Section~\ref{sec:relatedwork} reviews the works that combine early layers from ConvNets and the state-of-the-art methods in HAR associated with wearable data. Next, Section~\ref{sec:proposedmethod} explains our proposed LHN method. Finally, Section~\ref{sec:experiments} and~\ref{sec:conclusions} present our experimental results and concluding remarks, respectively.
\section{Related Work}\label{sec:relatedwork}
The use of early layers to improve the data representation has been explored in the context of object detection. However, these approaches demand a high computational cost\footnote{{The computation cost increases since a larger number of parameters must be learned by the ConvNet due to
		addition of $1\times1$ convolution and deconvolution layers.}} since the combination of the layers is performed by a deconvolution or $1\times1$ convolution layer, which makes it usage unfeasible in wearable systems

Bell et al.~\cite{Bell:2016} proposed a method to extract contextual and multi-scale information. For this purpose, their method, called Inside-Outside Net, combines previous layers from ConvNet. This combination is performed by normalizing and presenting the features maps to a $1\times1$ convolution, {a trick employed to avoid the high dimensionality}. The authors noticed that this combination aids, mainly, in the detection of small objects, which require higher spatial resolution produced by lower-level layers. Instead of combining the layers using a $1\times1$ convolution, as in~\cite{Bell:2016}, Kong et al.~\cite{Kong:2016} used max-pooling and deconvolution layers to re-scale all feature maps to the same size.  Then, the authors presented these re-scaled features maps to a fully-connected layer.

The idea behind using $1\times1$ convolution and deconvolution in~\cite{Bell:2016} and~\cite{Kong:2016} is to reduce the data dimensionality and compress the feature maps into a uniform space (to properly combine them), respectively. However, the number of parameters of the network increases considerably, since these new layers must be learned. In our proposed method, on the other hand, the number of parameters of the network is not modified (since we do not change the network) and the problem of the high dimensional of the data is easily handled by the PLS.

Previous works on human activity recognition based on wearable sensors  have showed that ConvNets approaches are able to provide significant improvements. Chen and Xue~\cite{Chen:2015} proposed a sophisticated ConvNet to classify the different categories of activities from raw signal. Their proposed ConvNet consists of three convolutional layers, where each layer is followed by a $2\times1$ max-pooling layer. Similar to~\cite{Chen:2015}, Jiang and Yin~\cite{Jiang:2015} proposed a shallow ConvNet with only two layers. However, to improve the activity representation, Jiang and Yin~\cite{Jiang:2015} suggested a method, called signal image, in which once the signal image is generated, a discrete Fourier transform is applied to it, producing a novel matrix which is presented as the  ConvNet input. 

Different from~\cite{Chen:2015} and~\cite{Jiang:2015}, Ha et al.~\cite{Ha:2015} proposed a multi-modal ConvNet consisting of convolutional filters and max-pooling of sizes $3\times3$ and $5\times5$, respectively. The filters at the first layer are learned separately to each heterogeneous modality (e.g., accelerometer and gyroscope). For this purpose the authors introduced zero-padding between the different modalities, in this way, the modalities are not merged during the convolution process. Following~\cite{Ha:2015}, Ha and Choi~\cite{Ha:2016} introduced zero-padding at the second layer to separate the filters of each modality in all layers from ConvNet. An interesting aspect of their work is that the authors demonstrated that ConvNets ($2$D convolutions) are more suitable to HAR {based on wearable sensors} than $1$D convolutions. They showed that the models generated by ConvNets are smaller and have, nearly, $4.7$ times fewer parameters than the $1$D convolutions, which is an important issue since mobile devices have limited memory and computational power. 

A drawback in~\cite{Ha:2015,Ha:2016}, however, is that due to their proposed ConvNet architecture (design of the filters), it is possible to execute these ConvNets only in datasets where multiple devices are used to capture the data (e.g., devices placed on different body parts to acquire accelerometer and gyroscope). {This restriction happens because the convolution process produces feature maps smaller than the input provided to it and its dimension can reach invalid values (i.e., zero) in deep ConvNets. Hence, in data provided by single devices the input samples are small, which contributes to the above-mentioned limitation}.

In contrast to the aforementioned studies, our work focuses on enhancing the data representation generated from ConvNets, which enables improvements independently of the sensor data and ConvNet architecture. This data representation improvement is achieved by multi-scale information (obtained through the combination of shallow with deep layers from ConvNet), which is enhanced by our LHN method. 
\section{Proposed Approach}\label{sec:proposedmethod}
In this section, we start by briefly describing the Partial Least Squares (PLS) {used for projecting the features maps in our method}. Afterwards, we introduce the approach to generate our proposed Latent Hypernet approach.

The PLS is a dimensionality reduction technique which projects the high dimensional space onto a latent space, where the covariance between the feature and its label is maximized. The PLS technique works as follows. Let $X \subset R^m$ be a matrix of {independent variables} representing the samples (activities) in $m$-dimensional space (originated by the layer from a ConvNet). Let $y$ be the matrix {dependent variables} denoting the class label in a $k$-dimensional space, where $k$ represents the number of categories of activities. The PLS projects $X$ onto a new $c$-dimensional space (where $c$ is the single parameter of the method), $X' \subset R^c$, in terms of $X' = XW$, where $W$ is a weight matrix and can be computed, iteratively, using the NIPALS algorithm~\cite{Wold:1985}, Algorithm~\ref{alg::nipals}.

The NIPALS algorithm computes a column of $W$ at each iteration (step $3$ in Algorithm~\ref{alg::nipals}), which represents the maximum covariance between $X$ and $y$. It should be mentioned that in Algorithm~\ref{alg::nipals}, the convergence step is achieved when, from an iteration to another, no change occurs in $w_a$ (defined in the algorithm). Additionally, we can use a fixed number of steps (i.e., $20$) to ensure the method stop in step $7$. Finally, before presenting the matrices $X$ and $y$ for the NIPALS algorithm they are normalized to operate in the same scale (transformed into $Z$-scores). Thereby, we ensure that the output of different layers works in a common scale.
\begin{algorithm}[!htb]
	\caption{NIPALS Algorithm.}
	\label{alg::nipals}
	\SetKwInOut{Input}{Input}
	\SetKwInOut{Output}{Ouput}
	\Input{$m$-dimensional data $X$, label matrix $y$}
	\Input{Number of components $c$}
	\Output{Weight matrix $W$}
	\BlankLine
	\For{ $a=1$ \textbf{to} $c$}{
		randomly initialize $u \in \mathbb{R}^{m \times 1}$
		\BlankLine
		$w_a = \frac{X^Tu}{\lVert X^Tu \rVert}$, where $w_a \in W$
		\BlankLine
		$t_a = Xw_a$
		\BlankLine
		$q_a = \frac{y^Tt_a}{\lVert y^Tt_a\rVert}$
		\BlankLine
		$u = yq_a$
		\BlankLine
		Repeat steps $3-6$ until convergence
		\BlankLine
		$p_a = X^Tt_a$
		\BlankLine
		$X = X - t_ap_a^T$
		\BlankLine
		$y = y - t_aq_a^T$
	}
\end{algorithm}

Note that other dimensionality reduction methods, such as principal component analysis or linear discriminant analysis, could be used to find the projection matrix $W$, however, this work employs PLS since many studies showed that it robust to unbalanced and multiclass problems~\cite{Schwartz:2009, Santos:2015, Kloss:2017}.

The generation of the LHN works as follows. First, we present all the training samples to the network and after each max-pooling\footnote{{We have selected the max-pooling layer since it is robust to spatial shift.}} layer $i$, we use its feature maps to learn a PLS model. Clearly, we can notice that each max-pooling layer will have a PLS model associated with it. Even though we could have concatenated the features provided by all max-pooling layers and then project the concatenated features to the latent space, generating a single PLS model, the memory consumption would increase significantly since the result of this concatenation is a high dimensional space and wearable devices have limited memory. Therefore, we perform the projection iteratively, layer by layer. In addition, iteratively projecting the layer enables the method to be efficient in deeper networks. Finally, we concatenate and present to a classifier all the latent features (LF) produced by each max-pooling layer i-$th$. Algorithm~\ref{alg::LHN} presents the steps of the described  process.

\begin{algorithm}[!htb]
	\caption{Latent HyperNet Algorithm.}
	\label{alg::LHN}
	\SetKwInOut{Input}{Input}
	\SetKwInOut{Output}{Output}
	\Input{ConvNet, $LF = \{\}$ }
	\Output{Concatenated Latent Features (LF)}
	\BlankLine
	\BlankLine
	\ForEach{max-pooling layer $\in ConvNet$}{
		$X_{i} =\text{features maps from max-pooling}_{i}$
		\BlankLine
		Find $W_{i}$ using Algorithm~\ref{alg::nipals}
		\BlankLine
		$X' = X_{i}W_{i}$ (Projection step in Figure~\ref{fig:LatentHyperNet})
		\BlankLine
		$LF = LF \cup X'$ (Concatenation step in Figure~\ref{fig:LatentHyperNet})
	}
\end{algorithm}
\section{Experimental Results}\label{sec:experiments}
{In this section, we first present the datasets employed to validate our proposed LHN approach. Then, we describe the experimental setup and our proposed ConvNets, respectively. Finally, we show the improvements achieved by LHN, the importance of the stage of dimensionality reduction, the computational cost of the method and compare our approach with other state-of-the-art methods.}

\vspace{2mm}
\noindent\textbf{Datasets.}
{Instead of describing each dataset individually, we summarize the main features of them in Table~\ref{tab::datasets}. From this table, it is possible to observe that the datasets vary in terms of the sampling rate (Freq.), number of available sensors and activities. In this way, we can examine the robustness of the methods regarding the high variation in the essence of the data.}
\begin{table}[!htb]
	\centering
	\caption{{Main features of each dataset. The available sensors in each dataset vary from Accelerometer (Acc) to Gyroscope (Gyro), and Magnetometer (Mag).}}
	\label{tab::datasets}
	\begin{tabular}{|c|c|c|c|}
		\hline
		Dataset                               & Freq. (Hz) & $\#$Sensors            & $\#$Activities \\ \hline
		USCHAD~\cite{Zhang:2012}       & $100$           & $2$ (Acc and Gyro)   & $12$         \\ \hline
		WISDM~\cite{Lockhart:2011}     & $20$            & $1$ (Acc)            & $7$          \\ \hline
		MHEALTH~\cite{Banos:2014}      & $50$            & $3$ (Acc, Gyro, Mag) & $12$         \\ \hline
		UTD-MHAD1~\cite{ChenChen:2015} & $50$            & $2$  (Acc, Gyro)     & $21$         \\ \hline
		UTD-MHAD2~\cite{ChenChen:2015} & $50$            & $2$ (Acc, Gyro)      & $5$          \\ \hline
	\end{tabular}
\end{table}

\vspace{2mm}
\noindent\textbf{Experimental Setup.}
Throughout the experiments, we adopt the $10$-fold cross-validation protocol, which is a standard protocol applied to HAR based on sensors~\cite{Zeng:2014, Shoaib:2015, Quiroz:2017}. To report the results, we use the recall metric~\cite{Powers:2011}, which we also refer to as recognition rate. 

{The input samples to the ConvNets are temporal windows generated from raw signal. These windows are produced by dividing the signal into subparts (windows) and considering each subpart as an entire activity. Formally, we can define a temporal window in terms of}
\begin{equation}
\label{eq::temporal_sliding_window}
\begin{split}
w =  [s_{k-t},..., s_{k-2}, s_{k-1}, s_k ]^\top,
\end{split}
\end{equation}
{where $k$ denotes the current sample captured by the sensor and $t$ denotes the temporal sliding window size. The windows that do not fit within the temporal window are dropped. For a detailed discussion regarding this procedure we recommend~\cite{Jiang:2015,Song:2017}.} In addition, following~\cite{Morris:2014, Song:2017}, we segment the raw signal using temporal sliding window of $5$ seconds (value $t$ in Equation~\ref{eq::temporal_sliding_window}). However, since the activities of UTD-MHAD~\cite{ChenChen:2015} dataset have the duration of 2-3 seconds, to this dataset, we use a window of 1 second, which limits the use of deeper architectures and large convolutional kernels. This happens because the convolution process generates feature maps smaller than the input provided to it and its size can reach zero in deep ConvNets.
\begin{figure}[!b]
	\centering
	\includegraphics[scale=0.65]{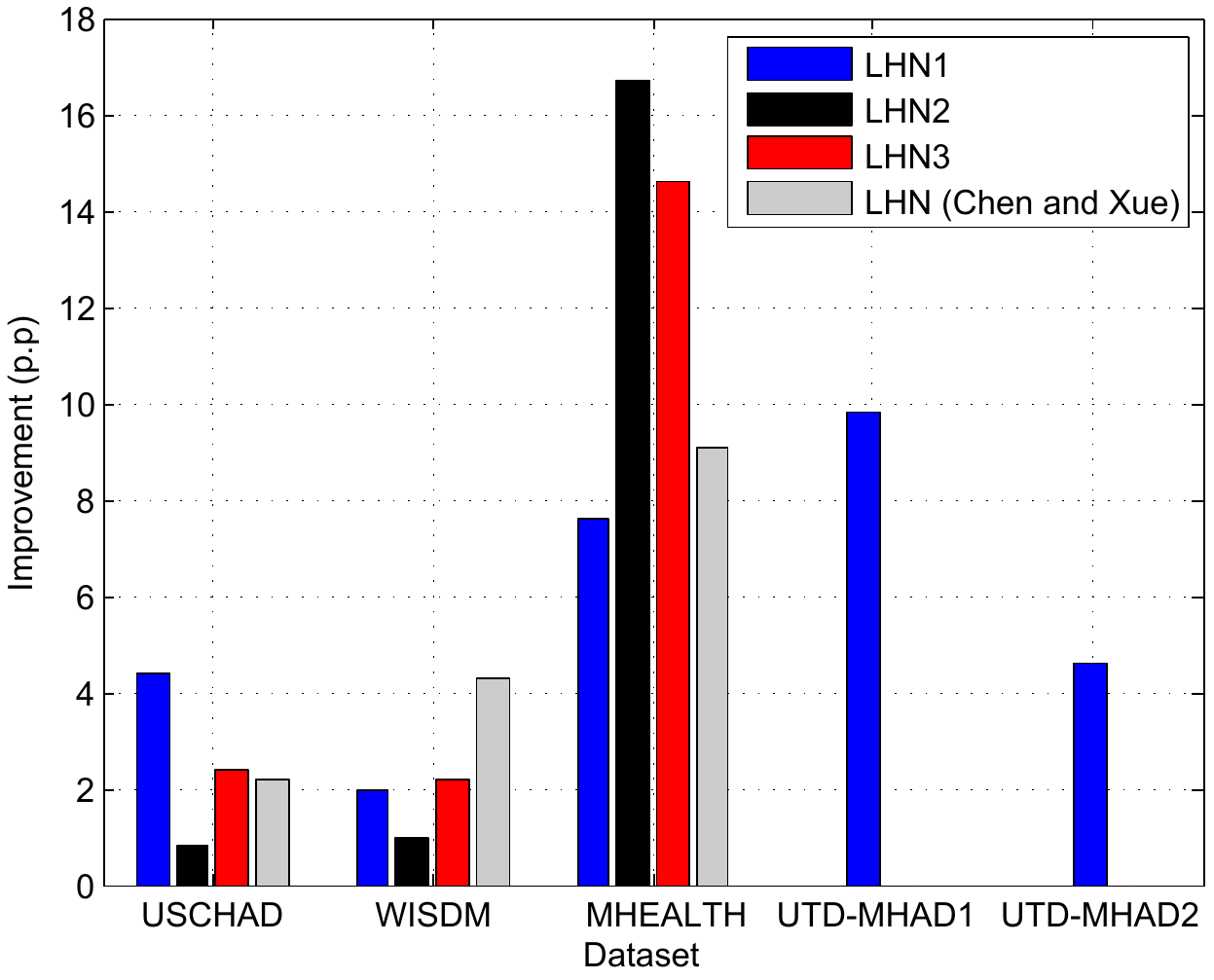}
	\caption{Improvements achieved by our LHN method compared to the original ConvNet (best viewed in color).}
	\label{fig:improvements}
\end{figure}
\begin{table*}[!htb]
	\centering
	\caption{Configurations of ours proposed ConvNets.}
	\label{tab:convnets}
	\begin{tabular}{|c|c|c|c|}
		\hline& \#Conv. Layers (Depth) & \# Filters per Layer & Kernel Shape (heigh $\times$ width) per Layer \\ \hline
		ConvNet1& $2$&$24, 32$&$12\times2, 12\times2$\\ \hline
		ConvNet2&$3$&$24, 32, 40$&$6\times1, 8\times1, 10\times1$         \\ \hline
		ConvNet3&$4$&$24, 32, 40, 48$&$12\times1, 12\times1, 6\times1, 2\times1$\\ \hline
	\end{tabular}
\end{table*}

\vspace{2mm}
\noindent\textbf{Convolutional Neural Networks.} As mentioned in the previous sections, to evaluate the LHN robustness regarding the ConvNets, we propose three different ConvNets (\emph{ConvNet1}, \emph{ConvNet2} and \emph{ConvNet3}), which vary in the number of filters, kernel dimensions (shape) and depth of the network. Table~\ref{tab:convnets} shows the architectures of our proposed ConvNets. The first column in this table shows the depth of the network (Figure~\ref{fig:LatentHyperNet} illustrates a ConvNet of depth $3$, i.e., three  convolutional layers). In particular, since each convolutional layers is followed by a max-pooling layer (with kernel of $2\times1$), the first column also indicates the number of max-pooling layers, i.e., the number of PLS models that compose the LHN.

\vspace{2mm}
\noindent\textbf{Latent HyperNet.}
Since the essence of the LHN is the dimensionality reduction step, we need to find the best number of components, $c$, to the PLS. For this purpose, we range $c$ from $1$ to $20$ and evaluate the results achieved using the USCHAD dataset~\cite{Zhang:2012}, where $c$ equals to $19$ yielded the best result. We use the same value on the other datasets. In addition, to render a fair comparison and show the improvement obtained by the LHN, we use the same classifier employed by the original ConvNet, which is a fully connected layer followed by a SoftMax classifier. In this way, our LHN is not biased by the classifier.

Figure~\ref{fig:improvements} shows the improvements (difference between the recall achieved by the ConvNet using our LHN and the one without using the LHN) achieved by the LHN method regarding the employed ConvNet, where the $ith$ LHN represents the LHN using the $ith$ ConvNet. The proposed LHN method was able to increase the activity recognition for all ConvNets. In particular, the LHN was able to improve up to $16.70$ percentage points (p.p.) the activity recognition, representing a significant improvement since many efforts have been done to achieve just minor improvements~\cite{Jiang:2015,Ha:2016}.

Considering all datasets evaluated in Figure~\ref{fig:improvements}, the proposed LHN was able to improve the ConvNets 1-3, on average, $5.68$ p.p., $6.16$ p.p. and $6.40$ p.p., respectively. Moreover, the use of the LHN enhanced the recognition rate in $6.20$ p.p. when employed to the architecture proposed by Chen and Xue~\cite{Chen:2015}. These results reinforce our hypothesis that shallower layers, when properly combined with deeper layers, are able to enhance the discrimination of activities, allowing a better activity recognition.

Although the achieved improvements seem small, many efforts have been done to achieve smaller improvements in HAR based on wearable sensor data. For instance, Catal et al.~\cite{Catal:2015} and Ha and Choi~\cite{Ha:2016} improved the works of Kwapisz et al.~\cite{Kwapisz:2010} and Ha et al.~\cite{Ha:2015} in $2.81$ p.p. and $2.19$ p.p., respectively. Therefore, our LHN achieves notable enhancements.

\vspace{2mm}
\noindent\textbf{Importance of the dimensionality reduction.} In this experiment, we show the importance of the dimensionality reduction step in our LHN method. To this end, we measure the results of the LHN without the dimensionality reduction step on the USC-HAD dataset~\cite{Zhang:2012}.
\begin{figure}[!b]
	\centering
	\includegraphics[scale=0.65]{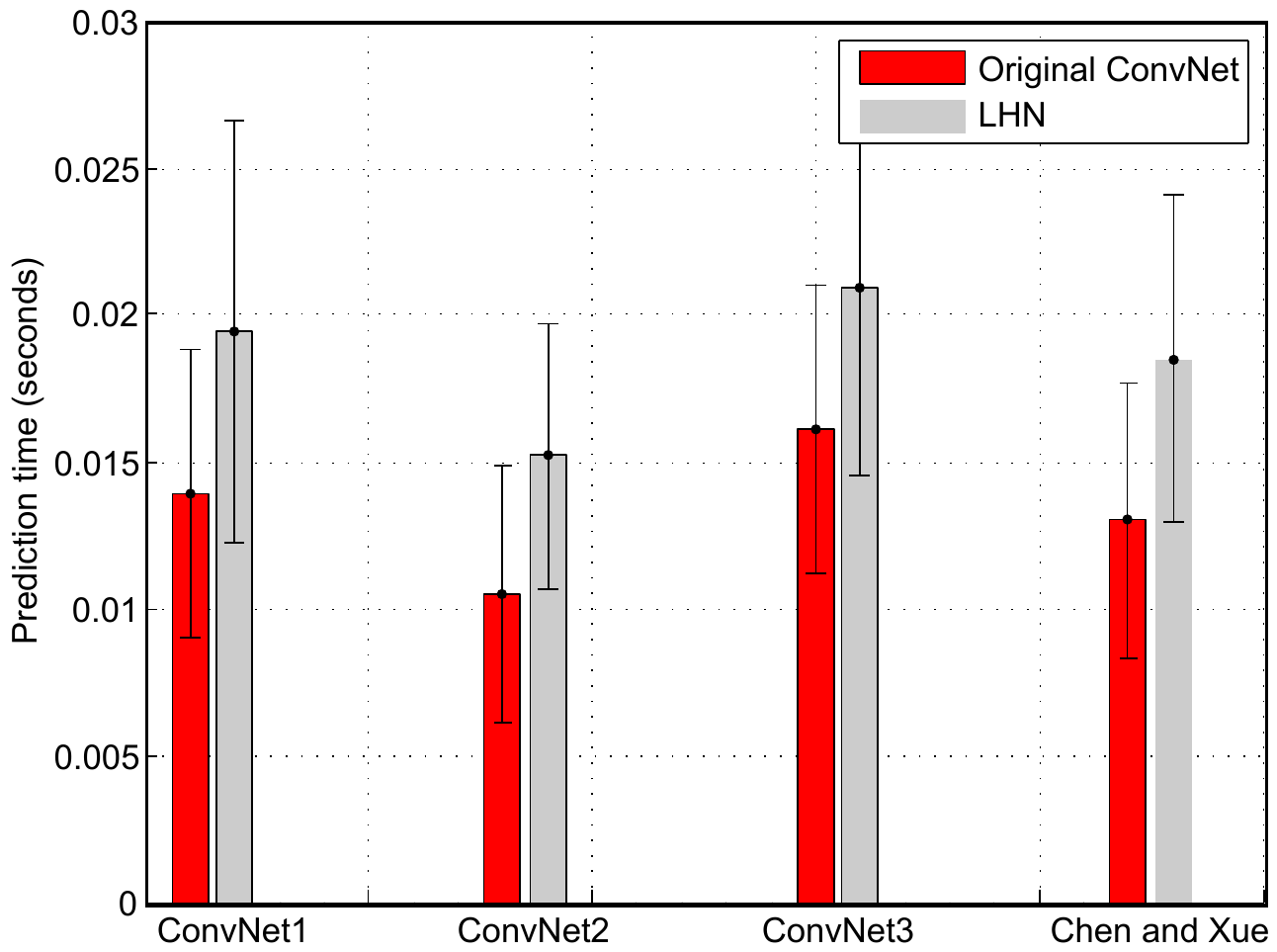}
	\caption{{Average prediction time, lower values are better (best viewed in color). It is possible to note that, our proposed method does not compromise the prediction time, since the time of the LHN is statistically equal to the original ConvNet time.}}
	\label{fig:time}
\end{figure}

By removing the dimensionality reduction, the recognition rate decreased $30$ p.p. on average. This occurs due to the high dimensionality generated from the concatenation of the feature maps, rendering the learning stage more complex since the network needs to learn a larger number of parameters. On the contrary, by using the dimensionality reduction we generate a low-dimensional feature space, where there are fewer parameters to be learned, which aids the learning stage.
\begin{table*}[!t]
	\centering
	\caption{Comparison with state-of-the-art methods. Values in bold denote the top $2$ best methods for each dataset. {ConvNet1-3 denote the proposed ConvNets without the employment of the LHN method. The last row indicates the ConvNet3 using the LHN method, except for the datasets UTD-MHAD1-2, where we use the LHN with the ConvNet2.} Cells with the symbol $-$ denote that it is not possible to execute the ConvNet on the respective dataset, due to its architecture.}
	\label{tab:state_of_the_art}
	\begin{tabular}{|c|c|c|c|c|c|}
		\hline &MHEALTH~\cite{Banos:2014}&USCHAD~\cite{Zhang:2012}&UTD-MHAD1~\cite{ChenChen:2015}&UTD-MHAD2~\cite{ChenChen:2015}&WISDM~\cite{Lockhart:2011} \\ \hline
		Jiang and Yin~\cite{Jiang:2015}& $55.6$& $76.4$& $\mathbf{42.0}$& $70.0$& $82.3$ \\ \hline
		Chen and Xue~\cite{Chen:2015}& $65.7$& $78.7$& -& -& $86.0$ \\ \hline
		Ha et al.~\cite{Ha:2015}& $67.9$& -& -& -& -\\ \hline
		Ha and Choi~\cite{Ha:2016}& $\mathbf{84.8}$& -& -& -& -\\ \hline
		\textbf{ConvNet1 (Ours)}& $68.2$& $80.7$& $40.3$& $\mathbf{70.7}$& $86.3$\\ \hline
		\textbf{ConvNet2 (Ours)}& $61.8$& $79.9$& - & - & $\mathbf{87.0}$\\ \hline
		\textbf{ConvNet3 (Ours)}& $63.5$& $\mathbf{81.4}$& -& -& $85.8$\\ \hline
		\textbf{LHN (Ours)}&$\mathbf{78.1}$ & $\mathbf{83.8}$ & $\mathbf{50.1}$ & $\mathbf{75.3}$ & $\mathbf{88.0}$  \\ \hline
	\end{tabular}
\end{table*}

\vspace{2mm}
\noindent\textbf{Time issues.}
{Since our method performs a projection after each max-pooling layer (as explained in Section~\ref{sec:proposedmethod}), it introduces an extra cost to predict the samples. In this experiment, we show that this cost is irrelevant, which enables the LHN method to be computationally efficient {compared to} the traditional ConvNet. To demonstrate that, we perform a statistical evaluation~\cite{Raj:1990}, on the prediction time. In this evaluation, we computed the confidence interval by using a confidence of $95\%$ and estimate the average prediction time by considering $30$ executions.
}

Figure~\ref{fig:time} shows the average and the confidence interval of the original ConvNets (red bars) and the ones using the proposed LHN method (gray bars). 
According to this figure, it is possible to observe that the confidence intervals present overlap, thereby, the methods might be statistically equivalent. To validate this claim, as suggested by Jain~\cite{Raj:1990}, we perform a \emph{unpaired t-test} between the methods. On this test, our method has been shown to be equivalent to original ConvNet, which makes the prediction time of the LHN statistically equivalent to the original ConvNet. Therefore, the employment of the LHN method does not compromise the prediction time.

\vspace{2mm}
\noindent\textbf{Comparison with the State-of-the-art.} As we mentioned earlier, the current state-of-the art results in HAR based on wearable sensors are achieved with methods based on ConvNets~\cite{Ha:2016,Wang:2017}. Therefore, our last experiment compares the LHN with such methods. It is important to note that all the methods used in this experiment are ConvNets dedicated to HAR based on wearable data. Moreover, to provide a fair comparison, we re-train all the methods on the same conditions (e.g., number of epochs and training samples). {Finally, we do not compare our method with LSTMs-based approaches since, while they presented good results in natural language processing~\cite{Greff:2017} and speech recognition~\cite{Graves:2013}, in the context of HAR based on wearable sensors ConvNets have presented superior results~\cite{Ha:2016}.}

According to Table~\ref{tab:state_of_the_art}, in most of the cases, our proposed ConvNets (even without using LHN) outperform existing ConvNets in the literature. We believe that our ConvNets provide better results due to the  convolutional kernel dimensions. For instance,~\cite{Jiang:2015,Ha:2015} use kernels of $3\times3$ and $5\times5$, respectively, which capture a small temporal pattern besides being sensitive to noise by data acquisition. On the other hand, our kernels are able to capture a large temporal relation of the signal and, hence, to be more robust to noise.

In order to compare our LHN with the state-of-the-art methods, we select the LHN using our proposed ConvNet3. However, since UTD-MHAD dataset does not enable deep architectures (as we argued before), we select the LHN using ConvNet$1$ for this dataset. Table~\ref{tab:state_of_the_art} shows that our proposed method outperforms state-of-the-art methods in activity recognition based on wearable data. {For instance, on the datasets USCHAD, UTD-MHAD2 and WISDM we outperformed the state-the-art in $5.1$ p.p., $5.3$ p.p. and $2.0$ p.p., respectively.} Moreover, it is important to notice that  the recognition rate was reduced drastically on the UTD-MHAD1 dataset due to the large number of activities contained in this dataset. However, our method outperformed the previous best method in $8.1$ p.p., demonstrating that our LHN provides a richer data representation. Finally, as can be noticed in Table~\ref{tab:state_of_the_art}, on the MHEALTH dataset the best recognition rate was achieved by the method of Ha and Choi~\cite{Ha:2016}. Additionally, when we apply the LHN method on this ConvNet, we obtain an improvement of $7.3$ p.p., outperforming the state-of-the-art once again.
\section{Conclusions}\label{sec:conclusions}
This work presented a robust and accurate method, referred to as Latent HyperNet (LHN), to improve ConvNets applied to HAR based on wearable sensor data. The method individually projects the features from each layer onto a latent space, where a richer representation of these features is obtained. We evaluate the proposed method using different ConvNet architectures and our experiments demonstrated that our method improves the recognition rate regarding the original ConvNet, {without compromising its the computational cost at the prediction stage, besides outperforming} existing state-of-the art methods. {We highlight that many efforts have been done to achieve small improvements in human activity recognition based on wearable sensor data, which reinforces that the LHN produces notable improvements.}

Since LHN does not modify the design of the ConvNet, it can be easily adaptable to any network. Therefore, as future work, we intend to apply it to other applications which employ ConvNets, such as image classification. In addition, we plan to employ other dimensionality reduction techniques, such as Linear Discriminant Analysis and Principal Components Analysis, to build the LHNs.
\section*{Acknowledgments}
The authors would like to thank the Brazilian National Research Council -- CNPq (Grant \#311053/2016-5), the Minas Gerais Research Foundation -- FAPEMIG (Grants APQ-00567-14 and PPM-00540-17) and the Coordination for the Improvement of Higher Education Personnel -- CAPES (DeepEyes Project).

\IEEEpeerreviewmaketitle

\balance
\bibliographystyle{IEEEtran}
\bibliography{References}

\end{document}